\def\year{2021}\relax
\documentclass[letterpaper]{article} 
\usepackage{aaai21}  
\usepackage{times}  
\usepackage{helvet} 
\usepackage{courier}  
\usepackage[hyphens]{url}  
\usepackage{graphicx} 
\usepackage{natbib}  
\usepackage{caption} 
\usepackage{subcaption}
\usepackage{amsmath}
\usepackage{amssymb}
\usepackage{algorithm}
\usepackage{algorithmic}
\usepackage{booktabs}

\title{Deep Stochastic Volatility Model}
\author {

        Xiuqin Xu,\textsuperscript{\rm 1}
        Ying Chen \textsuperscript{\rm 2}\\
}
\affiliations {
    \textsuperscript{\rm 1} Institute of Data Science and NUS Graduate School for Integrative Sciences
    \& Engineering, National University of Singapore \\
    \textsuperscript{\rm 2} Department of Mathematics and Risk Management Institute, National University of Singapore\\
    xiuqin.xu@u.nus.edu, matcheny@nus.edu.sg
}

\begin{document}

\maketitle

\begin{abstract}
Volatility for financial assets returns can be used to gauge the risk for financial market. 
We propose a deep stochastic volatility model (DSVM) based on the framework of deep latent variable models. 
It uses  flexible deep learning models to automatically detect the dependence of the future volatility on past returns, past volatilities and the stochastic noise, and thus provides a flexible volatility model without the need to manually select features. 
 We develop a scalable inference and learning algorithm based on  variational inference. 
 In real data analysis,  the DSVM outperforms several popular alternative volatility models.  
 In addition, the predicted volatility of the DSVM provides a more reliable risk measure that can better reflex the risk in the financial market, reaching more quickly to a higher level when the market becomes more risky and  to a lower level when the market is more stable, compared with the commonly used GARCH type model with a huge data set on the U.S. stock market.
\end{abstract}

\section{Introduction}
\label{sec:introduction}
The volatility of the financial assets returns, as a risk measure for the financial market, is very crucial for investment decision \citep{markowitz1952portfolio}.
It is also the most important factor in derivatives pricing  \citep{Duan1995,Goard2013}, since the uncertainty of the assets returns, reflected as volatility, drives the values of the derivatives. 
The volatility of assets returns is not observable and the historical volatility is usually measured by the standard deviation of the historical returns over a fixed time interval. 

The returns of financial assets often exhibit heteroscedasticity, in the sense that the volatility of the returns is time-dependent.
In particular, higher volatility tends to be followed by higher volatility, while lower volatility usually is followed by lower volatility, one phenomenon called volatility clustering in the financial market. 
In addition, the past positive and negative returns have asymmetric effects on the volatility --- a negative past return will increase the volatility more significantly than a positive past return.

In order to do future investment decision and derivatives pricing, volatility models need to be developed to forecast volatility in the future.
As the volatility is not observable, the biggest challenge in constructing volatility model is to identify the dependence between the unobserved volatility and the observed data (assets returns) so as to capture the above two empirical characteristics of the volatility: volatility clustering and asymmetric effects of positive and negative returns. 

Generalized Autoregressive Conditional Heteroscedasticity model (GARCH) \citep{bollerslev1986generalized} is the most popular volatility model, in which the variance of the returns (squared volatility) is assumed to be a linear function of the past squared returns and past squared volatility.
A number of variant GARCH models have been developed, most of which mainly attempt to address the two limitations of the GARCH model: the assumption of a linear relationship between current and past squared volatility and the assumption of symmetric effects of the positive and negative past returns.
Maximum likelihood method can be used to learn the parameters of the GARCH family models. 
A detailed review of the GARCH family can be find in \cite{hentschel1995all}. 
The GARCH family models are deterministic volatility models, in the sense that given the past returns and past volatility, the future volatility is a deterministic value. 
An alternative type of volatility models is the stochastic volatility model, in which the volatility is modeled as a latent stochastic process defined through partial differential equations.
In practice,  stochastic volatility models discretized for even spaced data are usually used instead for forecasting \citep{SKim1997,Omori2007}, and are estimated using Monte Carlo Markov Chain methods.
Although theoretically more sound, the stochastic volatility models usually do not outperform the GARCH model in practice \citep{SKim1997,Omori2007}.

Both the above deterministic and stochastic volatility models developed in statistics community impose restricted form on the function of the volatility model. 
Recently, volatility modeling has received attention from the machine learning community, where researchers are interested in proposing volatility model with less assumptions on the function form.
For example, \cite{Wu2014} propose to model the dependence of future volatility on past volatility and past return with a flexible non-linear Gaussian process model, which only assumes the volatility model function to be smooth. 
Recent developments in deep latent neural network models \citep{krishnan2017structured,fraccaro2016sequential,chung2015recurrent,bayer2014learning,Liu2018} provide a flexible framework to model the dependence between observed and unobservable latent variables.
\cite{Luo2018} propose a volatility model in which a sequence of latent variables are used to extract a compact representation of the original multivariate stock return time series. However, the model does not consider the effects of volatility clustering explicitly, but implicitly through the latent variables. In addition, modeling multivariate stock time series (using more information) entails more effects on data pre-processing, for example, how to choose the group of stocks to be modeled together, which may affect the performance.

In this work, we propose a Deep Stochastic Volatility Model (DSVM) based on the framework of the deep latent variable model using only the information of univariate return time series. The DSVM imposes no restriction on the function form for the volatility model, and explicitly considers the effects of stochastic noise,  past volatility (autoregression) and past returns (exogenous variables) on volatility, thus providing a general framework for volatility modeling. A scalable learning and inference method is developed to estimate the DSVM based on variational inference. Real data analysis shows that the DSVM  outperforms several popular volatility models in terms of testing negative likelihood based on only the information of the univariate return time series. In addition, the predicted volatility of DSVM gives a more reliable risk measure for financial assets: higher volatility when market is more risky and lower volatility when market is more stable.

In Section 2, we will review several volatility models in detail. The formulation of the DSVM is given in Section 3 and a scalable inference and learning algorithm is presented in Section 4. Section 5 concludes.

\section{Volatility models} \label{sec:volatility}
In this section, we will detail on several volatility models. Considering a sequence of T observations of the stock returns, labeled $r_{1:T}=\{r_1,r_2,\cdots,r_T\}$, we are interested in modeling the distribution $p(r_{1:T})$ as well as the latent volatility process $\sigma_t, t = 1,\cdots, T$.  One common approach is to assume the return $r_t$ follows a normal distribution with mean 0 and variance $\sigma_t^2$. Mathematically, we can formulate as
\begin{equation*}
\begin{aligned}
r_{t} &= \sigma_t \epsilon_t \\
\epsilon_t &\sim \text { i.i.d. }  \mathcal{N} \left(0, 1\right) 
\end{aligned}
\end{equation*}
Volatility models aim to find a function form for the evolution of the volatility $\sigma_t$ or some transformations of the volatility such as the variance $\sigma_t^2$ and the log variance $\log (\sigma_t^2)$. 

\subsection{Deterministic volatility models}
The GARCH \citep{bollerslev1986generalized} assumes a linear dependence of the future variance $\sigma_t^2$ on the previous $p$ squared returns and previous $q$  variances, computed as as follows:
\begin{equation*}
\sigma_{t}^{2} =\omega_{0}+\sum_{i=1}^{p} \alpha_{i} r_{t-i}^{2}+\sum_{j=1}^{q} \beta_{j} \sigma_{t-j}^{2}
\end{equation*}
Volatility clustering is captured by the parameters $\alpha_{i}, i=1,\cdots,p$. However, the GARCH model assumes symmetric effects of negative and positive returns. \cite{glosten1993relation} propose the GJR-GARCH model which extends the GARCH model to account for the asymmetry effects of positive and negative returns as follows:
\begin{equation*}
\sigma_{t}^{2}=\omega_{0} +\sum_{i=1}^{p} \alpha_{i} (r_{t-i} ^{2} + \gamma_{i} I\left\{r_{t-i}<0\right\}r_{t-i}^{2} )+\sum_{j=1}^{q} \beta_{j} \sigma_{t-j}^{2}
\end{equation*}
where the asymmetry effect is captured by the indicator function $I\left\{r_{t-i}<0\right\}$ and controlled by the leverage parameters $\gamma_{i}$. Similarly, the Threshold GARCH (TGARCH) proposed by \cite{zakoian1994threshold} also use leverage parameter $\gamma_{i}$ to reflect the asymmetry effects. The difference is that the TGARCH directly works with the volatility $\sigma_{t}$ instead of the variance  $\sigma_{t}^2$:
\begin{equation*}
\sigma_{t} =\omega_{0} +\sum_{i=1}^{p} \alpha_{i}(\left|r_{t-i}\right| - \gamma_{i}r_{t-i})+\sum_{j=1}^{q} \beta_{j} \sigma_{t-j} 
\end{equation*}
Another popular extension of the GARCH model is the Exponential GARCH (EGARCH, \cite{nelson1991conditional}), which models the evolution of the log variance $\log (\sigma_t^2)$:
\begin{equation*}
\log \left(\sigma_{t}^{2}\right) =\omega_{0}+\sum_{i=1}^{p} (\alpha_{i} \epsilon_{t}+\gamma_i\left(\left|\epsilon_{t}\right|-\mathbb{E}\left|\epsilon_{t}\right|\right))+\sum_{j=1}^{q} \beta_{j} \log \left(\sigma_{t-j}^{2}\right) 
\end{equation*}
where the leverage parameter $\gamma_i$ is used to account for the asymmetry effect.

\subsection{Stochastic volatility models}

For analysis convenience, stochastic volatility models are usually formulated with stochastic differential equations in continuous-time \citep{hull1987pricing,heston1993closed}. The Heston model for a univariate process is
\begin{equation*}
\begin{aligned} 
\mathrm{d} \sigma(t) &=-\beta \sigma(t) \mathrm{d} t+\delta \mathrm{d} W^{\sigma}(t) \\ \mathrm{d} s(t) &=\left(\mu-0.5 \sigma^{2}(t)\right) \mathrm{d} t+\sigma(t) \mathrm{d} W^{s}(t) 
\end{aligned}
\end{equation*}
where $s(t)$ is the logrithm of stock price at $t$, $W^{\sigma}(t)$ and  $W^{s}(t)$ are two correlated Wiener processes, with correlation represented as $E[\mathrm{d} W^{\sigma}(t)\mathrm{d} W^{s}(t)] = \rho \mathrm{d}t$. For practical use, stochastic volatility can be formulated in a discrete-time fashion for regularly spaced data. The discrete stochastic volatility without leverage \citep{SKim1997} is defined as
\begin{equation*}
\begin{aligned}
r_t &=  \sigma_t \epsilon_t\\
\log (\sigma_t^2) &=  \mu + \phi (\log (\sigma_{t-1}^2) -\mu ) + z_t\\
\epsilon_t \sim \text { i.i.d. } \mathcal{N}(0,1) &  \quad  z_t \sim \text { i.i.d. } \mathcal{N}(0,\sigma_z^2)
\end{aligned}
\end{equation*}
In order to accommodate the asymmetric effect of positive and negative returns, \cite{Omori2007} propose the discrete stochastic volatility with leverage as follows:
\begin{equation*}
\begin{array}{c}
{r_{t}=\sigma_{t} \epsilon_{t}} \\ 
{\log \left(\sigma_{t}^{2}\right)=\mu+\phi\left(\log \left(\sigma_{t-1}^{2}\right)-\mu\right)+z_{t}} \\
{\left(\begin{array}{c}{\epsilon_{t-1}} \\ {z_{t}}\end{array}\right) \sim \text { i.i.d. } \mathcal{N}_{2}(\mathbf{0}, \Sigma), \quad \Sigma=\left(\begin{array}{cc}{1} & {\rho \sigma_{z}} \\ {\rho \sigma_{z}} & {\sigma_{z}^{2}}\end{array}\right)}\end{array}
\end{equation*}
where the leverage $\rho <0$ is trying to capture the increase in volatility that follows a drop in the return.

\subsection{Machine learning stochastic volatility model}
\cite{Wu2014} proposed a Gaussian process volatility model (GP-Vol):
\begin{equation*}
\begin{aligned}
r_{t} &=\sigma_{t} \epsilon_{t}\\
\log(\sigma_t^2) &= f(\log(\sigma_{t-1}^2), r_{t-1}) + z_t\\
\epsilon_t \sim \text { i.i.d. } \mathcal{N}(0,1) &  \quad  z_t \sim \text { i.i.d. } \mathcal{N}(0,\sigma_z^2)
\end{aligned}
\end{equation*}
where $f$ can be modeled as a sample path of a Gaussian Process and $z_t$ is the stochastic noise. The model can be trained with Sequential Monte Carlo method.

The neural stochastic volatility model (NSVM) proposed by \citep{Luo2018} can be formulated as follows:
\begin{equation*}
\sigma_t = f(r_{1:t-1}, z_{1:t})
\end{equation*}
where $f$ is a deep neural network model. As will be detailed in the following, one of the main differences of our DSVM model and the NSVM is that we use a different generative probabilistic model to
account for the effect of past volatility and model the volatility as $\sigma_t = f_\sigma(\sigma_{1: t-1},r_{1:t-1}, z_{1:t})$. Another difference is that,  \citep{Luo2018} model the volatility utilizing the information from multiple stock return time series which are correlated to each other and thus utilize more information. However, how to properly group multiple stocks  is not a easy task in the first place. In this paper, we aim to propose a volatility model using less information, i.e. only based on the knowledge of the univariate stock return series.

Besides, there exist other machine learning volatility models using more exogenous variables, such as sentiment data \citep{xing2019sentiment} and using other advanced deep learning architectures, see \citep{zhang2018benchmarking}.

\section{Deep stochastic volatility model} \label{sec:dsvm}
In this section, we give the formulation for the Deep Stochastic Volatility  Model (DSVM). 

Recall that $r_t$ is the asset return and $\sigma_t$ is the volatility of the asset at timestep $t$. We introduce $z_t$ as the continuous stochastic noise to the volatility at timestep $t$. The stochastic noise is assumed to be normal distributed, whose parameters depend on past stochastic noise as follows:
\begin{equation*}
z _ { t } \sim N(m_t^{(p)}, (v_t^{(p)})^2)
\end{equation*}
\begin{equation*}
m_ { t }^{(p)}  = f _ { 1 } \left(  z  _ { t - 1 } \right) , \quad  v_ { t } ^{(p)} = f_ {2} \left( z  _ { t - 1 } \right)
\end{equation*}
where $z _ { t }$ is a Gaussian variable with parameter $m_t^{(p)},v_t^{(p)}$ given by MLP models $f_{1}$, $f_{2}$ that depend on $z _ { t - 1 }$.

In the DSVM, the volatility is assumed to depend on past returns, past volatilities and past stochastic noise  through a Recurrent Neural Network (RNN), computed as
\begin{equation*}
\begin{aligned}
h_t &= f_{h}(\sigma_{t-1},r_{t-1},z_{t},h_{t-1}) \\
\sigma_{t} &= f_3(h_t)
\end{aligned}
\label{equ:volatility}
\end{equation*}
where $f_h$ is a RNN model and $f_3$ is a MLP model. Therefore, current volatility depends on previous volatility $\sigma_{t-1}$, previous return $r_{t-1}$ and current stochastic noise $z_{t}$ as well as  $\sigma_{1:t-2}, r_{1:t-2}, z_{1:t-1}$ encoded in  $h_{t-1}$. This parameterization make  $\sigma_{t}$  depends on $\sigma_{1: t-1}, r_{1: t-1}, z_{1:t}$ and the volatility model can then be denoted as $\sigma_{t} = f_\sigma( \sigma_{1: t-1}, r_{1: t-1}, z_{1: t})$. 

The return  is assumed to follow a Gaussian distribution with mean 0 and standard deviation $\sigma_t$:
\begin{equation*}
r_{t} \sim \mathcal{N}(0,\sigma_t^2)
\end{equation*}

\begin{figure}[!t]
	\centering
	\includegraphics[width=0.6\linewidth]{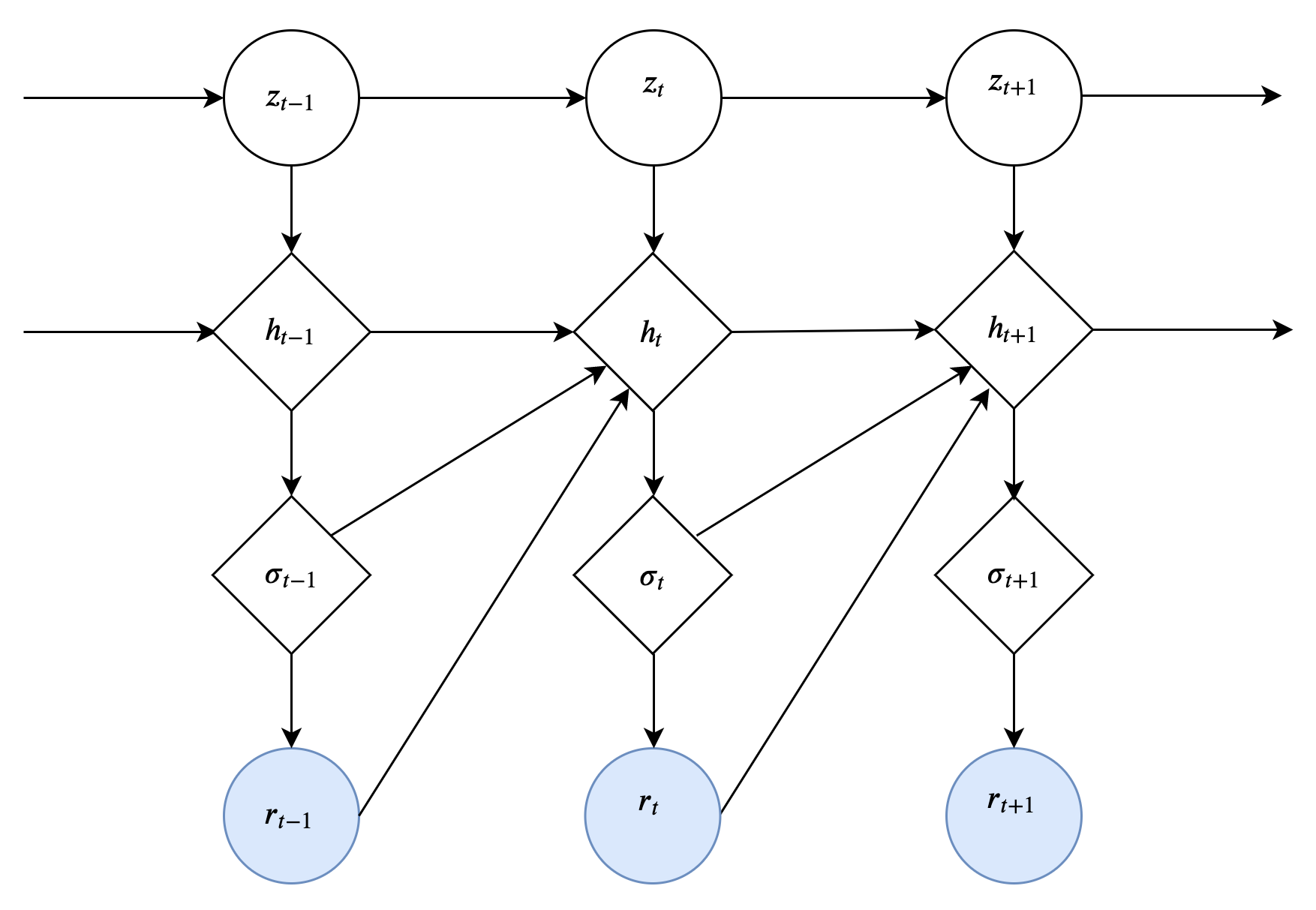}
	\caption{The generative network of the DSVM}
	\label{fig:deepvolatilitymodel}
\end{figure}

The graphical representation of the DSVM model is given in Figure \ref{fig:deepvolatilitymodel}, and is termed as the generative network which specifies the generating process of the data. The parameters of the generative network include the parameters in the following functions: $\theta =\{f_1,f_2,f_3,f_{h}\}$.

The joint probability of the return $r_{1 : T}$ and the stochastic noise $z_{1 : T}$ can be factorized as
\begin{align} 
&\quad p_{\theta}\left(r_{1 : T}, z_{1 : T} |\sigma_0=0, r_0=0,z_0=0\right) \notag\\
&=p_{\theta_{}}\left(r_{1 : T} | z_{1 : T}\right)p_{\theta_{}}\left(z_{1 : T} \right) \notag\\ 
&=\prod_{t=1}^{T} p_{\theta}\left(r_{t} | r_{1:t-1},  {z}_{1:t} \right)  p_{\theta}\left(z_{t} | z_{1:t-1}\right) \notag\\
&=\prod_{t=1}^{T} p_{\theta}\left(r_{t} |  \sigma_t \right)  p_{\theta}\left(z_{t} | z_{t-1}\right) \notag\\
&=\prod_{t=1}^{T} p_{\theta}\left(r_{t} |  \sigma_t  \right)  p_{\theta}\left(z_{t} | z_{t-1}\right).
\label{equ:jointprobability}
\end{align}

The marginal loglikelihood of the observed returns $r_{1 : T}$, denoted as $\mathcal{L}(\theta) = \log p_{\theta}\left(r_{1 : T}\right)$, can be obtained by averaging out  $z_{1 : T}$ in the joint distribution in Equation (\ref{equ:jointprobability}). However, integrating out $z_{1 : T}$ is challenging due to the non-linearility introduced by deep neural networks,  resulting intractable $\mathcal{L}(\theta)$. Therefore, maximum likelihood methods can not be used to estimate the parameters of DSVM model. In the next section, we will develop a scalable learning and inference for the DSVM using variational inference.

\section{Scalable learning and inference} \label{sec:learning}
Instead of maximizing the likelihood $\mathcal{L}(\theta)$ with respect to the parameter $\theta$, we instead maximize a variational evidence lower bound $ELBO(\theta,\phi)$ with respect to both $\theta$ and $\phi$, where $\phi$ is the parameter for the approximated posterior.

\subsection{Variational inference for the DSVM}
Denote $q_{\phi}\left(z_{1 : T} | r_{1 : T}\right)$ as any approximated posterior for the true posterior $p_{\theta}\left(z_{1 : T}| r_{1 : T}\right)$, it can be derived that
\begin{equation*}
\begin{aligned}
\mathcal{L}(\theta) &\geq ELBO(\theta,\phi) \\
&= \iint q_{\phi}\left(z_{1 : T} | r_{1 : T}\right) \log \frac{p_{\theta}\left(r_{1 : T}, z_{1 : T}  \right)}{q_{\phi}\left(z_{1 : T} | r_{1 : T}\right)} \mathrm{d} z_{1 : T}
\end{aligned}
\end{equation*}
When $q_{\phi}\left(z_{1 : T} | r_{1 : T}\right)$ is equal to the true posterior $p_{\theta}\left(z_{1 : T}| r_{1 : T}\right)$, the lower bound is tight and we have $\mathcal{L}(\theta) = ELBO(\theta,\phi)$. However, the true posterior is intractable.  In order to make the lower bound $ELBO$ tight, we design the inference network which represents $q_{\phi}\left(z_{1 : T} | r_{1 : T}\right)$ to best approximate the true posterior $p_{\theta}\left(z_{1 : T}| r_{1 : T}\right)$.
In fact, the true posterior can be factorized as
\begin{equation*}
\begin{aligned}
p_{\theta}\left(z_{1 : T}| r_{1 : T}\right) &= \Pi_{t=1}^Tp_{\theta}\left(z_{t} | {z}_{1 : t-1},r_{1 : T}\right) \\
&= \Pi_{t=1}^T p_{\theta}\left(z_{t}|z_{t-1}, r_{t:T}\right)
\end{aligned}
\end{equation*}

\begin{figure}[!t]
	\centering
	\includegraphics[width=0.6\linewidth]{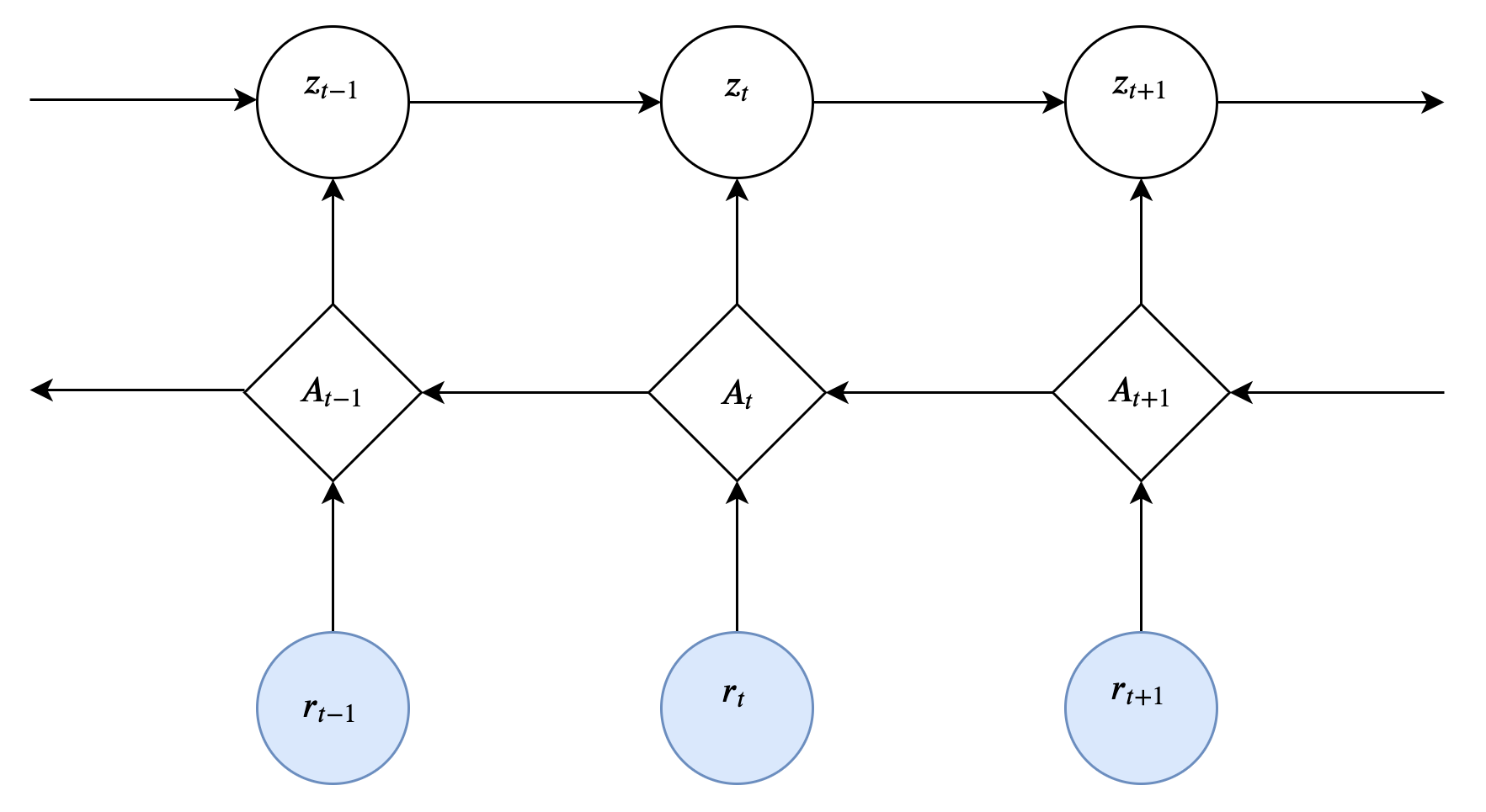}
	\caption{The inference network of the DSVM}
	\label{fig:inferencemodel}
\end{figure}

The factorization can be obtained from the Figure \ref{fig:deepvolatilitymodel} using d-seperation \citep{geiger1990identifying}. The factorization shows that, the posterior of ${z_t}$ depends on the past information encoded in $z_{t-1}$ as well as the future information in ${r}_{t : T}$. We can design our inference network according to this factorization form of the true posterior. Specifically,
\begin{equation}
q_{\phi}\left(z_{1 : T}| r_{1 : T}\right)= \prod_{t=1}^{T}q_{\phi}\left(z_{t} | {z}_{ t-1},{r}_{t : T}\right) 
\label{equ:inference}
\end{equation}

In addition, we have the following two factorization
\begin{equation*}
p_{\theta}\left(z_{1 : T}\right)  =\prod_{t=1}^{T} p_{\theta}\left(z_{t} | z_{t-1}\right)
\end{equation*}
\begin{equation*}
p_{\theta}\left(r_{1 : T} | z_{1 : T}\right) = \prod_{t=1}^{T} p_{\theta}\left(r_{t} | r_{1:t-1},  {z}_{1:t} \right) = \prod_{t=1}^{T} p_{\theta}\left(r_{t} |\sigma_t \right) 
\end{equation*}
The past information $r_{1:t-1},  {z}_{1:t}$ is reflected in $\sigma_t$. Therefore, the conditional density of return at time $t$ on the past returns and the stochastic noises can be replaced by the conditional density of return on its simultaneous volatility.

The ELBO then can be derived as
\begin{equation}
\resizebox{1\hsize}{!}{$
\begin{aligned}
& \quad \mathcal{L}(\theta) \\
&  \geq ELBO(\theta,\phi)\\
&= \int q_{\phi}\left(z_{1 : T} | r_{1 : T}\right) \log \frac{p_{\theta}\left(r_{1 : T}| z_{1 : T}\right) p_{\theta}\left(z_{1 : T}\right)}{q_{\phi}\left(z_{1 : T} | r_{1 : T}\right)} \mathrm{d} z_{1 : T} \\
&= \int q_{\phi}\left(z_{1 : T} | r_{1 : T}\right) \log p_{\theta}\left(r_{1 : T}| z_{1 : T}\right) \mathrm{d} z_{1 : T}  + \int q_{\phi}\left(z_{1 : T} | r_{1 : T}\right) \log \frac{p_{\theta}\left(z_{1 : T}\right)}{q_{\phi}\left(z_{1 : T} | r_{1 : T}\right)} \mathrm{d} z_{1 : T} \\
&=\mathbb{E}_{q_{\phi}\left(z_{1 : T}|r_{1 : T}\right)}\left[\log \prod_{t=1}^{T} p_{\theta}\left(r_{t} |  \sigma_t  \right) \right] -\int q_{\phi}\left(z_{1 : T} | r_{1 : T}\right) \log \frac{\prod_{t=1}^{T}q_{\phi}\left(z_{t} | z_{t-1},{r}_{t : T}\right) }{\prod_{t=1}^{T} p_{\theta}\left(z_{t} | z_{t-1}\right)}\mathrm{d} z_{1 : T}\\
&= \sum_{t=1}^{T}\mathbb{E}_{q_{\phi}^*(z_{t})}\left[\log p_{\theta}\left(r_{t} |  \sigma_t  \right) \right] -  \sum_{t=1}^{T} \mathbb{E}_{q_{\phi}^*(z_{t-1})} \left[KL[q_{\phi}\left(z_{t} | z_{t-1},{r}_{t : T}\right)|| p_{\theta}\left(z_{t} | z_{t-1}\right)]\right]
\label{equ:elbo_factorized}
\end{aligned}
$}
\end{equation}
where $q_{\phi}^*({z}_t)$ is the marginal probability density of ${z}_t$:
\begin{equation*}
\begin{aligned}
q_{\phi}^{*}\left({z}_{t}\right) &=\int q_{\phi}\left(z_{1 : T} |r_{1 : T}\right) \mathrm{d} {z}_{1 : t-1} \\
&= E_{q_{\phi}^{*}\left(z_{t-1}\right)}\left[q_{\phi}\left({z}_{t}|z_{t-1},{r}_{t : T}\right)\right]
\end{aligned}
\end{equation*}
We approximate the ELBO in Equation (\ref{equ:elbo_factorized}) using the Monte Carlo estimation. Specifically, we samples from $q_{\phi}^{*}\left({z}_{t}\right)$ using ancestral sampling according to the Equation (\ref{equ:inference}). Given a sample $z_{t-1}^{(s)}$ from $q_{\phi}^{*}\left(z_{t-1}\right)$, a sample ${z}_{t}^{(s)}$ from $q_{\phi}\left(z_{t} |z_{t-1}^{(s)},{r}_{t : T}\right)$ follows the distribution of $q_{\phi}^{*}\left({z}_{t}\right)$. After  we get the samples $z_t^{(s)} \text{ for } t= 1 \cdots, T$ from the $q_{\phi}\left(z_{1 : T}| r_{1 : T}\right)$, the ELBO in (\ref{equ:elbo_factorized})  can be approximated with the generated samples as:
\begin{align}
ELBO(\theta,\phi) &= \sum_{t=1}^{T}\left[\log p_{\theta}\left(r_{t} | \sigma_t = f_\sigma(\sigma_{1:t-1}, r_{1:t-1},  {z}_{1:t}^{(s)}) \right) \right] \nonumber \\
& \quad -  \sum_{t=1}^{T} \left[KL[q_{\phi}\left(z_{t} | z_{t-1}^{(s)},{r}_{t : T}\right)|| p_{\theta}\left(z_{t} | z_{t-1}^{(s)}\right)]\right]
\label{equ:approximatedELBO}
\end{align}

\subsection{Parameterization of the inference network}
Recall that, instead of factorizing $q_{\phi}$ as mean-field approximation  across different time step, i.e. $z_t, t = 1,\cdots, T$ are assumed to independent and does not use the information of the dependence structure in the generative network, we choose the variational posterior to have the same factorization as the true posterior in Equation \eqref{equ:inference}, where the posterior of $z_t$ depend on $z_{t-1}$ and $r_{t:T}$. 

We additionally assume that the information in $r_{t:T}$ can be encoded with a backward RNN, denoted as $g_A$ with hidden state, denoted as $A_t$,  as follows:
\begin{equation*}
\begin{aligned}
{A}_{t} &=g_{A}\left({A}_{t+1},r_{t}\right)\\
q_{\phi}\left(z_{1 : T}| r_{1 : T}\right) &= \prod_{t=1}^{T}q_{\phi}\left(z_{t} | {z}_{ t-1},{r}_{t : T}\right)  = \prod_{t=1}^{T}q_{\phi}\left(z_{t} | {z}_{ t-1}, A_t\right)
\end{aligned}
\end{equation*}
where $g_{A}$ is a  RNN model. The graphical model of the inference network is shown in Figure \ref{fig:inferencemodel}. $z_{t-1}$ represents the information coming from the past, whereas $A_t$ encodes the information  from present and the future, thus the information from all time steps are used to approximate the posterior at each $t$. Further, $q_{\phi_z}\left(z_{t} | z_{t-1}, A_t\right)$ is parameterized to be a Gaussian distribution density:
\begin{equation*}
z _ { t }  \sim N(m^{(q)}_t, (v^{(q)}_t)^2)
\end{equation*}
\begin{equation*}
m ^{(q)}_ { t }  = g _ { 1 } \left(  z  _ { t - 1 } ,  A  _ { t } \right) , \quad  v^{(q)} _ { t }  = g_ {2} \left( z  _ { t - 1 } , A  _ { t } \right)
\end{equation*}
$z _ { t }$ is  a Gaussian variable with parameter $m^{(q)}_t, v^{(q)}_t$ given by MLP models $g_{1}$ and $g_{2}$ that depend on $z _ { t - 1 } , A_t $. The parameters in the inference network include parameters in the following functions: $\phi =\{g_1,g_2,g_{A}\}$.

\subsection{The gradients}
The derivative of the $ELBO(\theta,\phi)$ with respect to $\theta$ can be calculated as:
\begin{equation*}
\begin{aligned}
&\quad \nabla_\theta ELBO(\theta,\phi)\\
&=  E_{q_{\phi}\left(z_{1 : T} | r_{1 : T}\right)} \nabla_\theta \log p_{\theta}\left(r_{1 : T}, z_{1 : T} \right)\\
&= \sum_{t=1}^{T} E_{q_{\phi}\left(z_{1 : T} | r_{1 : T}\right)} \nabla_\theta
[\log p_{\theta}\left(r_{t} | \sigma_t \right)  p_{\theta}\left(z_{t} | z_{t-1}\right)]
\end{aligned}
\end{equation*}
The derivative of the $ELBO(\theta,\phi)$ with respect to $\phi$ is more tricky as $\phi$ appears in the expectation in $ELBO(\theta,\phi)$. Score function gradient estimator \citep{glynn1987likelilood,williams1992simple} can be used to approximate the gradient, but the obtained gradients suffer from high variance. Reparameterization trick \citep{rezende2014stochastic,kingma2013auto}  is often used  to obtain a lower variance gradient estimator instead. Specifically, in order to obtain a sample $z_t$ from $ N(m_t^{(q)},(v_t^{(q)})^2)$, we first generate a sample $\eta_t \sim N(0,1)$, and then use  $m_t^{(q)} + \eta_t v_t^{(q)}$ as a sample of $z_t$, denote as $z_{t} = k_t(\eta_{t})$. The gradients then can be backpropogated through the gaussian random variables and $\nabla_\phi  ELBO(\theta,\phi)$ can be calculated as
\begin{equation*}
\begin{aligned}
&\quad \nabla_\phi ELBO(\theta,\phi) \\
&=\sum_{t=1}^{T} \nabla_\phi E_{q_{\phi}\left(z_{1 : T} | r_{1 : T}\right)} \log \frac{ p_{\theta}\left(r_{t} | \sigma_t \right)  p_{\theta}\left(z_{t} | z_{t-1}\right)}{q_{\phi}\left(z_{t} | {z}_{ t-1},{r}_{t : T}\right) }\\
&=  \sum_{t=1}^{T} \nabla_\phi E_{\eta_{1:t}\sim N(0,I)}  \log \frac{ p_{\theta}\left(r_{t} | \sigma_t \right)  p_{\theta}\left(k_t(
	\eta_{t}) | k_{t-1}(\eta_{t-1})\right)}{q_{\phi}\left(k_t(\eta_{t}) | k_{t-1}({\eta}_{ t-1}),{r}_{t : T}\right) }\\
&=  \sum_{t=1}^{T} E_{\eta_{1:t}\sim N(0,I)}  \nabla_\phi  \log \frac{ p_{\theta}\left(r_{t} | \sigma_t \right)  p_{\theta}\left(k_t(
	\eta_{t}) | k_{t-1}(\eta_{t-1})\right)}{q_{\phi}\left(k_t(\eta_{t}) | k_{t-1}({\eta}_{ t-1}),{r}_{t : T}\right) }
\end{aligned}
\end{equation*}

The above inference and learning algorithm for training DSVM are summarized in Algorithm \ref{alg:DSVM}.
\begin{algorithm}
	\caption{Structured Inference Algorithm for DSVM}
	\label{alg:DSVM}
	\begin{algorithmic}
		\STATE \textbf{Inputs}: $\{r_{1:T}\}_{i=1}^N$\\
		\qquad \qquad Randomly initialized $\phi^{(0)}$ and $\theta^{(0)}$\\
		\qquad \qquad Generative network model: $p_{\theta}\left(r_{1 : T}, z_{1 : T}\right)$\\
		\qquad \qquad Inference network model: $q_{\phi}\left(z_{1 : T} |r_{1 : T}\right)$
		\STATE \textbf{Outputs}: the parameters of $\theta$, $\phi$
		\WHILE{$Iter$ $<$ M} 
		\STATE1. Sample a mini-batch sequences $\{r_{1:T}\}_{i=1}^B$ uniformly from dataset
		\STATE2. Generate samples of $\eta_t^{(s)},t=1,\cdots,T$ to obtain samples of $z_t^{(s)},t=1,2,\cdots,T$ sequentially according to Equation (\ref{equ:inference}) to approximate the ELBO in Equation (\ref{equ:approximatedELBO})
		\STATE3. Derive the gradient $\nabla_\theta ELBO(\theta,\phi)$ which is approximated with one Monte Carlo sample
		\STATE4. Derive the gradient $\nabla_\phi ELBO(\theta,\phi)$  which is approximated with one Monte Carlo sample
		\STATE5. Update $\theta^{(Iter)}, \phi^{(Iter)}$ using the ADAM.
		\STATE set $Iter = Iter+1$
		
		\ENDWHILE
	\end{algorithmic}
\end{algorithm}

\subsection{One-step-ahead forecasting}
After training the model, we can use the estimated model to do one-step-ahead forecasting. Specifically, the probability of $r_{T+1}$ conditional on the information of $r_{1:T}$ can be derived as: 
\begin{equation}
\resizebox{1\hsize}{!}{$
\begin{aligned}
&\quad \quad p(r_{T+1}|r_{1:T}) \\
&= \int_{z_{1:T+1}} p(r_{T+1},z_{1:T+1}|r_{1:T}) d z_{1:T+1}\\
& =  \int_{z_{1:T+1}} p(r_{T+1}|z_{1:T+1},r_{1:T}) p(z_{1:T+1}|r_{1:T})d z_{1:T+1}\\
& =  \int_{z_{1:T+1}} p(r_{T+1}|\sigma_{T+1}= f_\sigma(\sigma_{1:T},r_{1:T}, z_{1:T+1})) p(z_{T+1}|z_{T})p(z_{1:T}|r_{1:T})d z_{1:T+1}\\
&\approx \int_{z_{1:T+1}} p(r_{T+1}|\sigma_{T+1}= f_\sigma(\sigma_{1:T},r_{1:T}, z_{1:T+1})) p(z_{T+1}|z_{T})q(z_{1:T}|r_{1:T})d z_{1:T+1}\\
&\approx \frac{1}{S}\sum_{s=1}^S p(r_{T+1}|\sigma_{T+1}= f_\sigma(\sigma_{1:T},r_{1:T}, z_{1:T+1}^{(s)}))
\end{aligned}
$}
\label{equ:predicted}
\end{equation}

We approximate the $p(z_{1:T}|r_{1:T})$ with  $q(z_{1:T}|r_{1:T})$ and use ancestral sampling to generate samples from the approximated $p(r_{T+1}|r_{1:t})$. Specifically, we generate  $\{z_{1:T}^{(s)}\}_{s=1}^{S}$ from $q(z_{1:T}|r_{1:T})$ and then generate samples $\{z_{T+1}^{(s)}\}_{s=1}^{S}$ from $p(z_{T+1}|z_{T})$ based on $\{z_{T}^{(s)}\}_{s=1}^{S}$. The approximated predictive distribution for $p(r_{T+1}|r_{1:t})$  is then a mixture of $S$ Gaussian models. We can then generate $\{r_{T+1}^{(s)}\}_{s=1}^{S}$ from the predictive distribution (\ref{equ:predicted}) and use the sample standard deviation of $\{r_{T+1}^{(s)}\}_{s=1}^{S}$ as the predicted volatility for $r_{T+1}$.

\section{Real data analysis} \label{sec:rda}
In this section, we present the experiment results of the proposed DSVM model on the US stock market.

\subsection{Data and model setting}
We choose the daily adjusted return of the 39 biggest stocks in terms of market values among SP500 stocks from 2001-01-02 to 2018-12-31.
For each stock, the training and validation data are from 2001-01-02 to 2015-10-18 (in total of 3721 days) and the testing data are from 2015-10-19 to 2018-12-31 (in total of 806 days). We extract sequences of 10 timestep data (corresponding to half of a trading month) from the original univariate time series and  the training, validation, testing data size is 105924, 31434 and 31473 respectively (a proportion of around 6:2:2).

We train one model for all the 39 stocks. The dimension of the latent variable $z$ is chosen to be 1. $f_1,f_2,f_3,g_1,g_2$ are all parameterized by two-hidden-layer MLP models with different activation functions at last layer (Linear for mean function $f_1$, $g_1$, Softplus for standard deviation function $f_2,f_3,g_2$). $f_h, g_{{A}}$ are both parameterized by one-layer GRU models with 10 hidden states. Monte Carlo samples size used for forecasting is 1000. The model is trained on a Tesla V100 GPU for 1 hours (300 epochs) before it converges. The final model is chosen by minimizing the validation loss over the 300 epochs. The DSVM is only trained once on the training and validation data, and then is used to do recursive one-day-ahead forecasting without retraining.

\subsection{Alternative models}
We compare the performance of the propsed model with seven popular alternative volatility models described in Section 2.
\begin{enumerate}
	\item Deterministic volatility models:\\
	We consider four models: GARCH, GJR-GARCH, TGARCH and EGARCH. The parameter estimation is done with a widely used R package ``rugarch" \footnote{https://cran.r-project.org/web/packages/rugarch}.
	\item Stochastic volatility models:\\
	We choose two models: the discrete stochastic volatility model without leverage and discrete stochastic volatility model with leverage, and use the R package ``stochvo"  \footnote{https://cran.r-project.org/web/packages/stochvol} to estimate the parameters.
	\item Gaussian process volatility model (GP-Vol) \\
	The implementation of the model is obtained from the authors' website\footnote{https://bitbucket.org/jmh233/code-gpvol-nips2014/src/master} and the hyperparameters are chosen as suggested in \cite{Wu2014}.
\end{enumerate}

For each type of the alternatives, we train one model for each stock as the alternatives are designed to be so. We will perform recursive one-day-ahead forecasting and all the alternative models  are retrained at every forecasting time step with a rolling window of 1000.

\subsection{Result and discussion}
We first evaluate each model in terms of negative loglikelihood (NLL) for the testing samples for each stock. The smaller the NLL value, the better the model can explain the distribution of the observed testing samples. 
The results of NLL evaluations are listed in Table \ref{tab:nll}. 
We first apply the Friedman rank sum test and the test indicates that there exist significant differences of the NLL values among different models (the p-value is less than $e^{-35}$). Furthermore, a pairwise pos thoc Nemenyi-Friedman test \citep{Demsar2006} is performed to compare different models, and it confirms that the DSVM performs best among all the methods. Our DSVM is the only stochastic volatility model that is able to beat deterministic volatility models and is the best for 36  out of the 38 stocks.


\begin{table}[!t]
	\footnotesize
	\centering
	\caption{Negative log-likelihood (NLL) evaluated on the test samples of 39 stock return time series for different volatility models. The best performing model are highlighted in bold. The last line shows the mean NLL for the 39 stocks. NA values for TGARCH are due to non-convergence of the parameter estimation (MLE).}
	{\resizebox{1.02\linewidth}{!}{
			\begin{tabular}{lrrrrrrrr}
				\toprule
				\toprule
				\multicolumn{1}{c}{\textbf{Stock}} & \multicolumn{1}{c}{\textbf{DSVM}} & \multicolumn{1}{c}{\textbf{GARCH}} & \multicolumn{1}{c}{\textbf{GJRGARCH}} & \multicolumn{1}{c}{\textbf{TGARCH}} & \multicolumn{1}{c}{\textbf{EGARCH}} & \multicolumn{1}{c}{\textbf{GPVol}} & \multicolumn{1}{c}{\textbf{StoVol}} & \multicolumn{1}{c}{\textbf{StoVolL}} \\
				\midrule
				\textbf{ORCL}  & \textbf{-3.03} & -2.87 & -2.87 & -2.87 & -2.88 & -2.82 & -2.25 & -2.23 \\
				\textbf{MSFT}  & \textbf{-2.96} & -2.84 & -2.86 & -2.86 & -2.87 & -2.87 & -2.44 & -2.45 \\
				\textbf{KO}    & -3.35 & -3.36 & -3.36 & -3.38 & \textbf{-3.38} & -3.27 & -3.14 & -3.15 \\
				\textbf{XOM}   & \textbf{-3.11} & -3.10 & -3.10 & \textbf{-3.11} & \textbf{-3.11} & -2.99 & -2.96 & -2.97 \\
				\textbf{GE}    & \textbf{-2.87} & -2.81 & -2.83 & -2.83 & -2.80 & -2.72 & -2.46 & -2.46 \\
				\textbf{IBM}   & \textbf{-3.06} & -2.92 & -2.92 & NA& -2.93 & -2.96 & -2.49 & -2.51 \\
				\textbf{PEP}   & -3.29 & \textbf{-3.30} & -3.30 & NA & -3.30 & -3.21 & -3.13 & -3.14 \\
				\textbf{MO}    & \textbf{-3.08} & -3.00 & -2.99 & -3.01 & -3.03 & -2.97 & -2.71 & -2.73 \\
				\textbf{COP}   & \textbf{-2.52} & -2.51 & \textbf{-2.52} & \textbf{-2.52} & \textbf{-2.52} & -2.32 & -2.40 & -2.42 \\
				\textbf{AMGN}  & \textbf{-2.92} & -2.85 & -2.86 & -2.86 & -2.86 & -2.81 & -2.67 & -2.70 \\
				\textbf{SLB}   & \textbf{-2.82} & \textbf{-2.82} & \textbf{-2.82} & \textbf{-2.82} & \textbf{-2.82} & -2.73 & -2.73 & -2.77 \\
				\textbf{CVX}   & \textbf{-2.95} & -2.93 & -2.94 & -2.93 & -2.93 & -2.80 & -2.80 & -2.84 \\
				\textbf{AAPL}  & \textbf{-2.89} & -2.82 & -2.85 & -2.85 & -2.85 & -2.77 & -2.52 & \textbf{-2.59} \\
				\textbf{UTX}   & \textbf{-3.16} & -3.12 & -3.13 & -3.13 & -3.12 & -3.05 & -2.85 & -2.89 \\
				\textbf{PG}    & \textbf{-3.27} & -3.25 & -3.25 & -3.26 & -3.26 & -3.18 & -2.94 & -2.98 \\
				\textbf{BMY}   & \textbf{-2.85} & -2.67 & -2.66 & NA & -2.67 & -2.71 & -2.07 & -2.08 \\
				\textbf{BA}    & \textbf{-2.86} & -2.77 & -2.77 & -2.76 & -2.76 & -2.71 & -2.50 & -2.50 \\
				\textbf{ABT}   & \textbf{-3.00} & -2.91 & -2.91 & -2.92 & -2.90 & -2.82 & -2.51 & -2.57 \\
				\textbf{PFE}   & \textbf{-3.18} & -3.15 & -3.16 & -3.15 & -3.15 & -3.07 & -2.96 & -2.97 \\
				\textbf{JNJ}   & \textbf{-3.27} & -3.20 & -3.21 & -3.19 & -3.19 & -3.16 & -2.83 & -2.83 \\
				\textbf{MMM}   & \textbf{-3.17} & -3.09 & -3.08 & -3.11 & -3.09 & -3.06 & -2.67 & -2.71 \\
				\textbf{MRK}   & \textbf{-3.08} & -2.99 & -3.01 & NA & -2.99 & -2.92 & -2.71 & -2.75 \\
				\textbf{DIS}   & \textbf{-3.12} & -3.07 & -3.07 & -3.08 & -3.08 & -3.04 & -2.86 & -2.88 \\
				\textbf{WFC}   & \textbf{-2.93} & -2.88 & -2.87 & -2.88 & -2.88 & -2.82 & -2.67 & -2.74 \\
				\textbf{MCD}   & \textbf{-3.20} & -3.07 & -3.04 & -3.06 & -3.13 & -3.03 & -2.71 & -2.75 \\
				\textbf{JPM}   & \textbf{-2.98} & -2.93 & -2.93 & -2.93 & -2.93 & -2.83 & -2.66 & -2.74 \\
				\textbf{WMT}   & \textbf{-3.09} & -2.90 & -2.87 & -2.94 & -2.94 & -2.88 & -2.34 & -2.34 \\
				\textbf{INTC}  & \textbf{-2.85} & -2.73 & -2.76 & -2.76 & -2.76 & -2.65 & -2.33 & -2.33 \\
				\textbf{BAC}   & \textbf{-2.74} & -2.71 & -2.71 & -2.71 & -2.71 & -2.61 & -2.53 & -2.58 \\
				\textbf{VZ}    & \textbf{-3.09} & -3.05 & -3.04 & -3.04 & -3.05 & -2.96 & -2.81 & -2.80 \\
				\textbf{T}     & \textbf{-3.09} & -3.03 & -3.03 & -3.04 & -3.04 & -2.91 & -2.66 & -2.65 \\
				\textbf{HD}    & \textbf{-3.12} & -3.10 & -3.11 & -3.11 & -3.11 & -3.02 & -2.94 & -2.98 \\
				\textbf{AIG}   & \textbf{-3.02} & -2.91 & -2.91 & -2.90 & -2.89 & -2.76 & -2.60 & -2.63 \\
				\textbf{C}     & \textbf{-2.82} & -2.78 & -2.79 & -2.80 & -2.80 & -2.71 & -2.58 & -2.65 \\
				\textbf{CSCO}  & \textbf{-3.01} & -2.91 & -2.93 & -2.95 & -2.95 & -2.86 & -2.50 & -2.56 \\
				\textbf{QCOM}  & \textbf{-2.72} & -2.52 & -2.52 & -2.50 & -2.53 & -2.45 & -1.51 & -1.50 \\
				\textbf{BRK}   & \textbf{-3.20} & -3.16 & -3.16 & -3.15 & -3.15 & -3.09 & -2.93 & -2.91 \\
				\textbf{AMZN}  & \textbf{-2.77} & -2.58 & -2.60 & NA & -2.65 & -2.69 & -2.24 & -2.29 \\
				\textbf{UNH}   & \textbf{-3.06} & -3.00 & -3.01 & -3.01 & -3.02 & -2.91 & -2.81 & -2.86 \\
				\midrule
				\textbf{Mean}  & \textbf{-3.01} & -2.94 & -2.94 & -2.95 & -2.95 & -2.88 & -2.63 & -2.65 \\
				\bottomrule
				\bottomrule
			\end{tabular}%
	}}
	\label{tab:nll}%
\end{table}%

\begin{figure}
	\centering
	\begin{subfigure}{1\linewidth}
		\centering
		\includegraphics[width=1\linewidth]{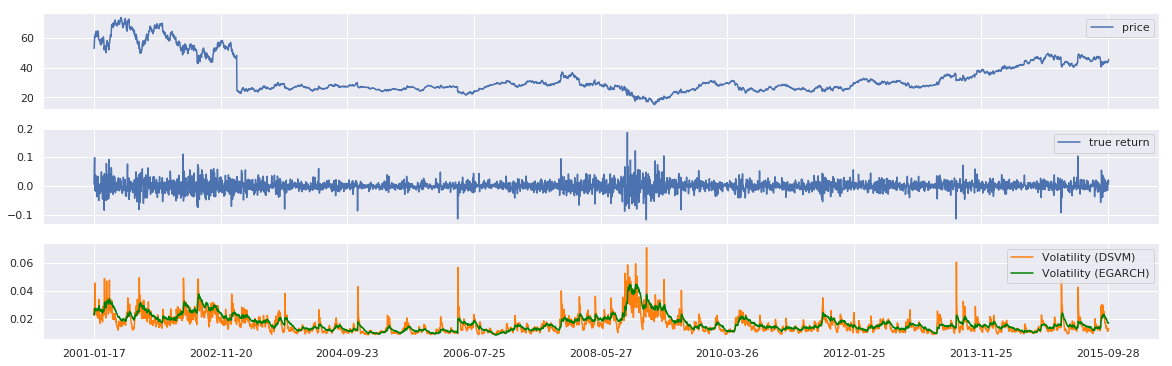}
		\subcaption{Stock MSFT }
		\label{fig:Stock1Insample}
	\end{subfigure}
	\begin{subfigure}{1\linewidth}
		\centering
		\includegraphics[width=1\linewidth]{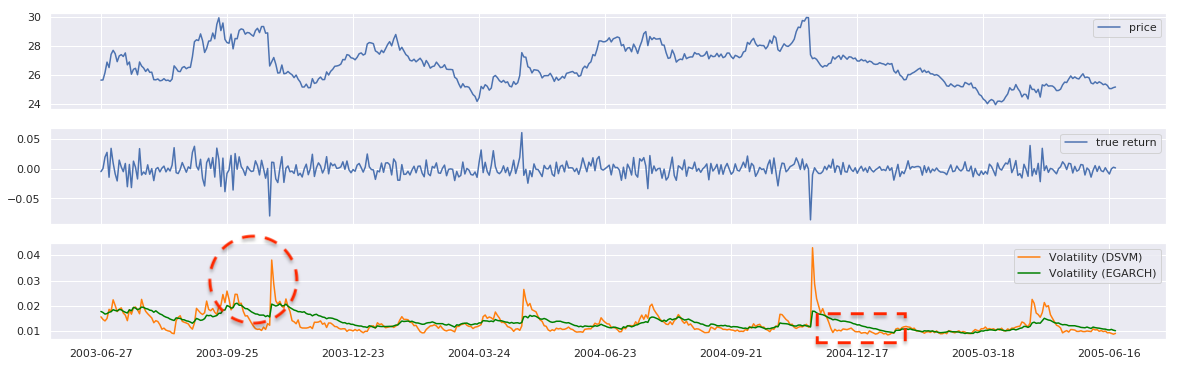}
		\subcaption{Stock MSFT, zoom in for a subsample }
		\label{fig:Stock1InsampleZoom}
	\end{subfigure}
	\begin{subfigure}{1\linewidth}
		\centering
		\includegraphics[width=1\linewidth]{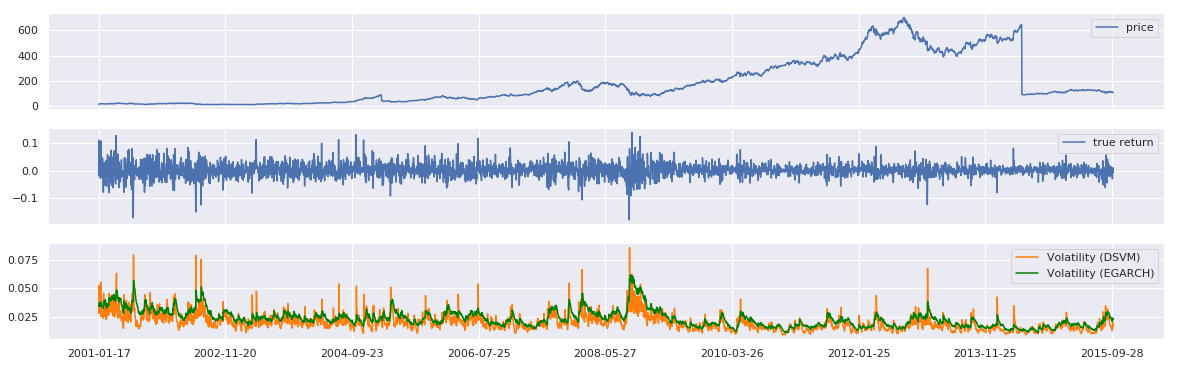}
		\subcaption{Stock AAPL}
		\label{fig:Stock2Insample}
	\end{subfigure}
	\begin{subfigure}{1\linewidth}
		\centering
		\includegraphics[width=1\linewidth]{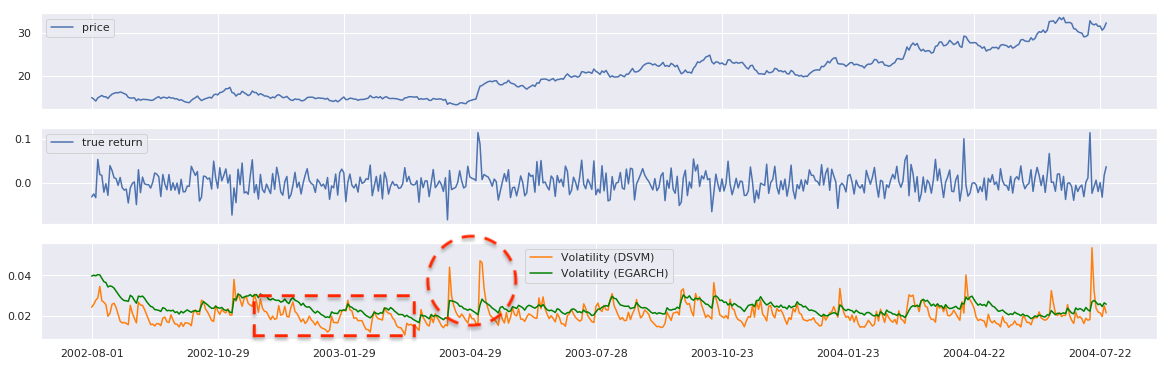}
		\subcaption{Stock AAPL, zoom in for a subsample }
		\label{fig:Stock2InsampleZoom}
	\end{subfigure}
	\caption{Training data}
	\label{fig:insample}
\end{figure}

\begin{figure}
	\begin{minipage}{1\linewidth}
		\centering
		\includegraphics[width=1\linewidth]{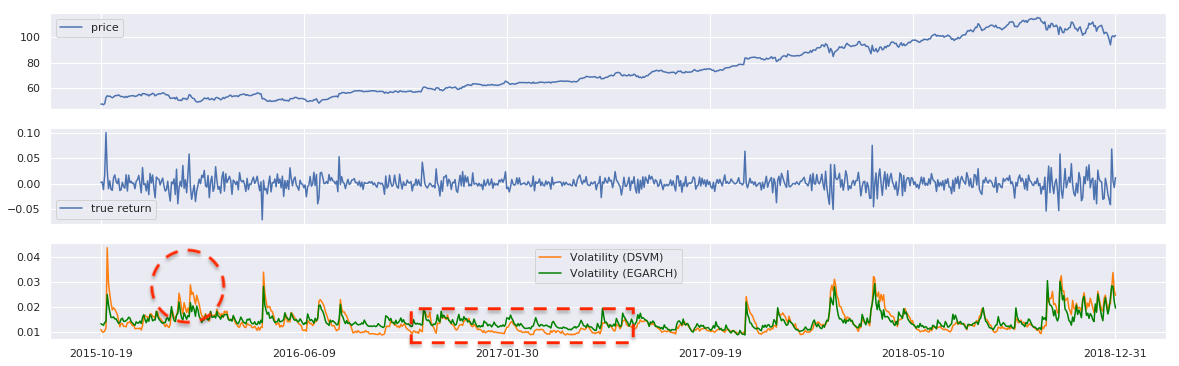}
		\subcaption{Stock MSFT}
		\label{fig:Stock1Outsample}
	\end{minipage}
	\begin{minipage}{1\linewidth}
		\centering
		\includegraphics[width=1\linewidth]{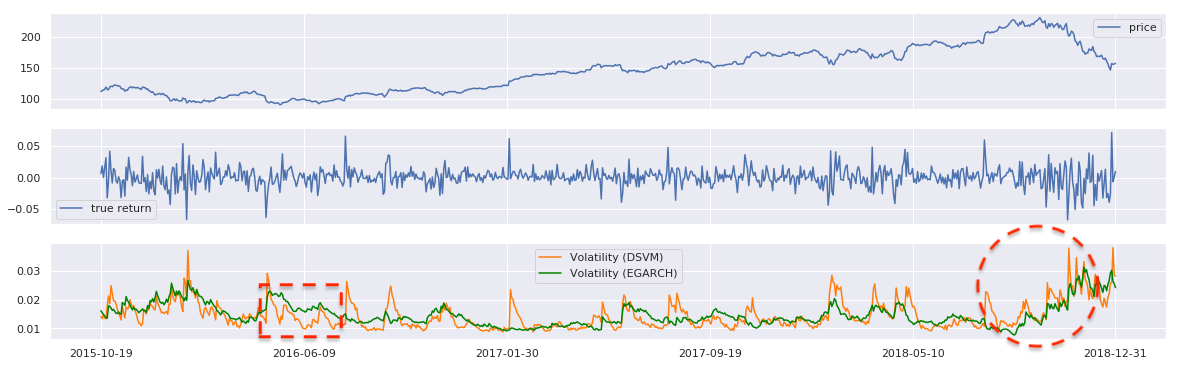}
		\subcaption{Stock AAPL }
		\label{fig:Stock2Outsample}
	\end{minipage}
	\caption{Testing data with recursive one-day-ahead forecasting}
	\label{fig:outsample}
\end{figure}

We further plot the predicted volatility of DSVM for training data and testing data and compare with the volatility produced by the EGARCH model (the best performing alternative model in terms of mean NLL). 
We use the two largest stocks (in terms of market value), the MSFT and the AAPL, for illustration. 
Other stocks perform similarly and are omitted in the manuscript due to page limit. 
The predicted volatility for training and testing data are displayed in Figure \ref{fig:insample} (with a randomly selected zoom-in period) and Figure \ref{fig:outsample} respectively. 
In general, the changes of the DSVM is more adaptive to the market than the EGARCH model, which is reflected by several spikes in the predicted volatility of DSVM.
In addition, the DSVM produces higher predicted volatility when the market is more volatile (see the circle areas) and gives lower volatility when the market is more stable (see the rectangle areas) for both training and testing data. 
In contrast, the predicted volatility of EGARCH is more smooth and the changes of the market status could not be quickly captured. 
Take the AAPL stock for example (see Figure \ref{fig:outsample}), the volatility of the EGARCH  decays more slowly around 2016 June when the market was calm, while it increases more slowly on 2018 September when the market became more volatile. 
In summary, the DSVM produces predicted volatility more responsive to the market status, which is more desirable as a risk measure for the financial market.

\section{Conclusion} \label{sec:conclusion}
In this work, we propose a deep stochastic volatility model based on the framework of deep latent variable models. 
The proposed volatility model does not impose restrictions on the function form on the volatility model, instead, it uses the flexible deep learning models to automatically detect the dependence of the future volatility on past returns, past volatilities and the stochastic noise. We developed a scalable inference and learning algorithm based on the variational inference, in which the parameters of the generative network and the inference network are jointly trained.
In addition, the results of real data analysis on the U.S. stock market show that the DSVM model using only the information from the univariate return time series outperforms several alternative volatility models. Furthermore, the
the predicted volatility of the deep stochastic volatility model is more adaptive to the market status, and thus provides a more reliable risk measure to better reflect the risk in the financial market.

\clearpage
\bibliographystyle{dcu}
\bibliography{refxx.bib}

\begin{thebibliography}{27}
\providecommand{\natexlab}[1]{#1}
\providecommand{\url}[1]{\texttt{#1}}
\providecommand{\urlprefix}{URL }
\expandafter\ifx\csname urlstyle\endcsname\relax
  \providecommand{\doi}[1]{doi:\discretionary{}{}{}#1}\else
  \providecommand{\doi}{doi:\discretionary{}{}{}\begingroup
  \urlstyle{rm}\Url}\fi

\bibitem[{Bayer and Osendorfer(2014)}]{bayer2014learning}
Bayer, J.; and Osendorfer, C. 2014.
\newblock Learning Stochastic Recurrent Networks.
\newblock In \emph{NIPS 2014 Workshop on Advances in Variational Inference}.

\bibitem[{Bollerslev(1986)}]{bollerslev1986generalized}
Bollerslev, T. 1986.
\newblock Generalized autoregressive conditional heteroskedasticity.
\newblock \emph{Journal of Econometrics} 31(3): 307--327.

\bibitem[{Chung et~al.(2015)Chung, Kastner, Dinh, Goel, Courville, and
  Bengio}]{chung2015recurrent}
Chung, J.; Kastner, K.; Dinh, L.; Goel, K.; Courville, A.~C.; and Bengio, Y.
  2015.
\newblock A recurrent latent variable model for sequential data.
\newblock In \emph{Advances in Neural Information Processing Systems},
  2980--2988.

\bibitem[{Dem{\v{s}}ar(2006)}]{Demsar2006}
Dem{\v{s}}ar, J. 2006.
\newblock {Statistical comparisons of classifiers over multiple data sets}.
\newblock \emph{Journal of Machine Learning Research} 7: 1--30.
\newblock ISSN 15337928.

\bibitem[{Duan(1995)}]{Duan1995}
Duan, J.~C. 1995.
\newblock {The GARCH option pricing model}.
\newblock \emph{Mathematical Finance} 5(1): 13--32.
\newblock ISSN 14679965.
\newblock \doi{10.1111/j.1467-9965.1995.tb00099.x}.

\bibitem[{Fraccaro et~al.(2016)Fraccaro, S{\o}nderby, Paquet, and
  Winther}]{fraccaro2016sequential}
Fraccaro, M.; S{\o}nderby, S.~K.; Paquet, U.; and Winther, O. 2016.
\newblock Sequential neural models with stochastic layers.
\newblock In \emph{Advances in Neural Information Processing Systems},
  2199--2207.

\bibitem[{Geiger, Verma, and Pearl(1990)}]{geiger1990identifying}
Geiger, D.; Verma, T.; and Pearl, J. 1990.
\newblock Identifying independence in Bayesian networks.
\newblock \emph{Networks} 20(5): 507--534.

\bibitem[{Glosten, Jagannathan, and Runkle(1993)}]{glosten1993relation}
Glosten, L.~R.; Jagannathan, R.; and Runkle, D.~E. 1993.
\newblock On the relation between the expected value and the volatility of the
  nominal excess return on stocks.
\newblock \emph{The Journal of Finance} 48(5): 1779--1801.

\bibitem[{Glynn(1987)}]{glynn1987likelilood}
Glynn, P.~W. 1987.
\newblock Likelilood ratio gradient estimation: an overview.
\newblock In \emph{Proceedings of the 19th Conference on Winter Simulation},
  366--375. ACM.

\bibitem[{Goard and Mazur(2013)}]{Goard2013}
Goard, J.; and Mazur, M. 2013.
\newblock {Stochastic volatility models and the pricing of vix options}.
\newblock \emph{Mathematical Finance} 23(3): 439--458.
\newblock ISSN 09601627.
\newblock \doi{10.1111/j.1467-9965.2011.00506.x}.

\bibitem[{Hentschel(1995)}]{hentschel1995all}
Hentschel, L. 1995.
\newblock All in the family nesting symmetric and asymmetric garch models.
\newblock \emph{Journal of Financial Economics} 39(1): 71--104.

\bibitem[{Heston(1993)}]{heston1993closed}
Heston, S.~L. 1993.
\newblock A closed-form solution for options with stochastic volatility with
  applications to bond and currency options.
\newblock \emph{The Review of Financial Studies} 6(2): 327--343.

\bibitem[{Hull and White(1987)}]{hull1987pricing}
Hull, J.; and White, A. 1987.
\newblock The pricing of options on assets with stochastic volatilities.
\newblock \emph{The Journal of Finance} 42(2): 281--300.

\bibitem[{Kingma and Welling(2013)}]{kingma2013auto}
Kingma, D.~P.; and Welling, M. 2013.
\newblock Auto-encoding variational bayes.
\newblock \emph{arXiv preprint arXiv:1312.6114} .

\bibitem[{Krishnan, Shalit, and Sontag(2017)}]{krishnan2017structured}
Krishnan, R.~G.; Shalit, U.; and Sontag, D. 2017.
\newblock Structured Inference Networks for Nonlinear State Space Models.
\newblock In \emph{AAAI}, 2101--2109.

\bibitem[{Liu et~al.(2018)Liu, He, Bai, Dai, Bai, and Xu}]{Liu2018}
Liu, H.; He, L.; Bai, H.; Dai, B.; Bai, K.; and Xu, Z. 2018.
\newblock {Structured inference for recurrent hidden semi-Markov model}.
\newblock \emph{IJCAI International Joint Conference on Artificial
  Intelligence} 2018-July: 2447--2453.
\newblock ISSN 10450823.

\bibitem[{Luo et~al.(2018)Luo, Zhang, Xu, and Wang}]{Luo2018}
Luo, R.; Zhang, W.; Xu, X.; and Wang, J. 2018.
\newblock {A neural stochastic volatility model}.
\newblock \emph{32nd AAAI Conference on Artificial Intelligence, AAAI 2018}
  6401--6408.

\bibitem[{Markowitz(1952)}]{markowitz1952portfolio}
Markowitz, H. 1952.
\newblock Portfolio selection.
\newblock \emph{The Journal of Finance} 7(1): 77--91.

\bibitem[{Nelson(1991)}]{nelson1991conditional}
Nelson, D.~B. 1991.
\newblock Conditional heteroskedasticity in asset returns: A new approach.
\newblock \emph{Econometrica: Journal of the Econometric Society} 347--370.

\bibitem[{Omori et~al.(2007)Omori, Chib, Shephard, and Nakajima}]{Omori2007}
Omori, Y.; Chib, S.; Shephard, N.; and Nakajima, J. 2007.
\newblock {Stochastic volatility with leverage: Fast and efficient likelihood
  inference}.
\newblock \emph{Journal of Econometrics} 140(2): 425--449.
\newblock ISSN 03044076.
\newblock \doi{10.1016/j.jeconom.2006.07.008}.

\bibitem[{Rezende, Mohamed, and Wierstra(2014)}]{rezende2014stochastic}
Rezende, D.~J.; Mohamed, S.; and Wierstra, D. 2014.
\newblock Stochastic Backpropagation and Approximate Inference in Deep
  Generative Models.
\newblock In \emph{International Conference on Machine Learning}, 1278--1286.

\bibitem[{{S. Kim}, {N. Sheppard}, and Chibb(1997)}]{SKim1997}
{S. Kim}; {N. Sheppard}; and Chibb, S. 1997.
\newblock {Stochastic Volatility: likelihood inference and comparison with ARCH
  models}.
\newblock \emph{Review of Economic Studies} 81: 159--192.

\bibitem[{Williams(1992)}]{williams1992simple}
Williams, R.~J. 1992.
\newblock Simple statistical gradient-following algorithms for connectionist
  reinforcement learning.
\newblock \emph{Machine Learning} 8(3-4): 229--256.

\bibitem[{Wu, Lobato, and Ghahramani(2014)}]{Wu2014}
Wu, Y.; Lobato, J. M.~H.; and Ghahramani, Z. 2014.
\newblock {Gaussian process volatility model}.
\newblock \emph{Advances in Neural Information Processing Systems} 1044--1052.
\newblock ISSN 10495258.

\bibitem[{Xing, Cambria, and Zhang(2019)}]{xing2019sentiment}
Xing, F.~Z.; Cambria, E.; and Zhang, Y. 2019.
\newblock Sentiment-aware volatility forecasting.
\newblock \emph{Knowledge-Based Systems} 176: 68--76.

\bibitem[{Zakoian(1994)}]{zakoian1994threshold}
Zakoian, J.-M. 1994.
\newblock Threshold heteroskedastic models.
\newblock \emph{Journal of Economic Dynamics and Control} 18(5): 931--955.

\bibitem[{Zhang et~al.(2018)Zhang, Luo, Yang, and Liu}]{zhang2018benchmarking}
Zhang, Q.; Luo, R.; Yang, Y.; and Liu, Y. 2018.
\newblock Benchmarking deep sequential models on volatility predictions for
  financial time series.
\newblock \emph{arXiv preprint arXiv:1811.03711} .

\end{thebibliography}

\end{document}